\documentclass[english]{lni}

\usepackage{mathtools}
\usepackage{xcolor}    
\usepackage{multirow}
\usepackage{graphicx}
\usepackage{fancyhdr}
\usepackage{changepage} 
\usepackage{url}

\usepackage[figurename=Fig., tablename=Tab., small]{caption}[2008/04/01]
\usepackage{enumitem}
\usepackage{subcaption}

\fancypagestyle{titlepage}{
\fancyfoot{}}

\renewcommand{\headrulewidth}{0.4pt} 
\author{Naser Damer$^{1 2}$ , Jonas Henry Grebe$^{1}$, Cong Chen$^{1}$, Fadi Boutros$^{1 2}$, Florian Kirchbuchner$^{1}$, Arjan Kuijper
\footnote{Fraunhofer Institute for Computer Graphics Research IGD, Darmstadt, Germany}\hspace{0.5mm} 
\footnote{Mathematical and Applied Visual Computing, TU Darmstadt, Darmstadt, Germany}}

\title{The Effect of Wearing a Mask on Face Recognition Performance: an Exploratory Study}
\begin{document}

\maketitle

\renewcommand{\refname}{References}
\setcounter{footnote}{6} 
\thispagestyle{titlepage}
\pagestyle{fancy}
\fancyhead{} 
\fancyhead[RO]{\small The Effect of Masks on Face Recognition \hspace{5pt} \thepage \hspace{0.05cm}}
\fancyhead[LE]{\hspace{0.05cm}\small \thepage \hspace{5pt} Naser Damer et al.}
\fancyfoot{} 
\renewcommand{\headrulewidth}{0.4pt} 

\begin{abstract}
Face recognition has become essential in our daily lives as a convenient and contactless method of accurate identity verification.
Process such as identity verification at automatic border control gates or the secure login to electronic devices are increasingly dependant on such technologies. 
The recent COVID-19 pandemic have increased the value of hygienic and contactless identity verification.
However, the pandemic led to the wide use of face masks, essential to keep the pandemic under control.
The effect of wearing a mask on face recognition in a collaborative environment is currently sensitive yet understudied issue.
We address that by presenting a specifically collected database containing three session, each with three different capture instructions, to simulate realistic use cases.
We further study the effect of masked face probes on the behaviour of three top-performing face recognition systems, two academic solutions and one commercial off-the-shelf (COTS) system.
\end{abstract}

\section{Introduction}


Given the current COVID-19 pandemic, it is essential to enable contactless and smooth running operations, especially in contact sensitive facilities like airports. Face recognition have been been praised as such an accurate and contactless mean of verifying identities. Wearing masks is essential to prevent the spread of contagious diseases and have been currently forced in public places in many countries. However, the performance, and thus the trust, of contactless identity verification through face recognition can be effected by wearing a mask.


Face occlusion have been repeatedly addressed in the scope of face detection solutions \cite{DBLP:conf/eccv/OpitzWPPB16}. Moreover, developing occlusion invariant face recognition solutions has been a growing research challenge \cite{DBLP:conf/iccv/SongGLLL19}. However, most of these works address general occlusion that commonly appear in in-the-wild capture conditions, such as sunglasses and partial captures. Given the current COVID-19 pandemic, it is essential to study the specific effect of wearing face masks on the behaviour of face recognition system in a collaborative environment.
Our work aims at studying this effect to enable the future development of solutions addressing accurate face recognition in such scenarios.
To achieve that, we present a database that simulates a realistically variant collaborative face capture scenario. This database is a first version of an on-going data collection process that includes three session, each with three capture variations, per subject. We study the behaviour of three of the top performing face recognition solutions (one commercial and two academic) when encountering masked faces, in comparison to the typical no-mask baseline. We conclude with pointing out strong signs of negative effect on face recognition systems, showing the need to develop appropriate evaluation databases and recognition solutions.

\section{Related work}

Face recognition deployment faces a number of operational challenges.
Many of these challenges, and thus the research efforts, are related to attacks on face recognition systems, such as presentation attacks (spoofing) \cite{DBLP:conf/bmvc/DamerD16}, morphing attacks \cite{DBLP:conf/icb/DamerSZWTKK19}, or other unconventional attacks \cite{DBLP:conf/btas/DamerWBBT0K18}.
However, issues related to the biometric sample presentation, such as face occlusion, can also effect face recognition deployability.
The detection of occluded faces is a well-studied issue in the computer vision domain. An example of that is the work of Optiz et al. \cite{DBLP:conf/eccv/OpitzWPPB16} that proposed a novel grid loss targeting a more accurate detection of occluded faces.
Focusing on masked faces, Ge et al. \cite{DBLP:conf/cvpr/GeLYL17} presented a solution to enhance the detection (not biometric recognition) of masked faces in in-the-wild scenarios. Their experiments did not focus on masks worn specifically for health protection reasons, but included other forms of face occlusions. However, their solution is relevant to face recognition as our experiments will show later that the investigated face recognition solutions fails in some cases to detect a face.


As stated, detecting occluded faces is a challenge that affect the operation of face biometric systems. However, the biometric recognition of these faces is a more dominant challenge. An example of the works addressing this challenge is that of Song et al. \cite{DBLP:conf/iccv/SongGLLL19} where they aim at enhancing face recognition for faces with general occlusions. Their approach tries to learn finding and discarding corrupted feature elements, linked to occlusions, from the recognition process.
Focusing on masks, in a very recent work, Wang et al. \cite{wang2020masked} presented, in a brief and undetailed work, crawled databases for face detection, recognition and simulated masked faces. The authors claim to enhance the recognition accuracy from 50\% to 95\% without providing information on their baseline, proposed algorithmic details, or clearly specifying the evaluation database. Given the current COVID-19 pandemic, a specifically collected database and evaluation of wearing real face mask on collaborative face recognition is necessary and is still missing.

\section{The database}

The goal of the collected database is to enable the study of face recognition performance on masked faces and drive future innovation in this domain. The database presented in this work is an initial version and further data collection efforts is on going. The data tries to simulate a collaborative, yet varying, scenario. Such as the situation in automatic border control gates or unlocking personal devices with face recognition, where the mask, illumination, and background can change.

Each of the participants was asked to collect the data on three different, not necessary consecutive days. We consider each of these days as one session. On each day, the participant will collect three videos, each of a minimum length of 5 seconds. All videos are collected from static (not hand held) webcams and the users were asked to simulate a login scenario by looking at the capture device. The images were all captured indoors, each at their residence during home-office. The capture was performed during the day (day-light) and the participants were asked to remove eyeglasses only when the frame is considered very thick. No other restrictions were imposed, such as background or mask type and its consistency over days, to simulate realistic scenarios. The three videos captured each day were as follows: 1) Face with no mask and no additional electric illumination, this will be noted as baseline (BL). 2) Face with mask on and no additional electric illumination, this will be noted as mask one (M1). 3) Face with mask on and the existing electric light in the room is turned on, this will be noted as mask two (M2). The M2 is considered to study the unknown effect of illumination variation in the case of masked face recognition, given that the mask might result in different reflection and shadow patterns.

\begin{figure}
     \centering
     \begin{subfigure}[b]{0.3\textwidth}
         \centering
         \includegraphics[width=\textwidth]{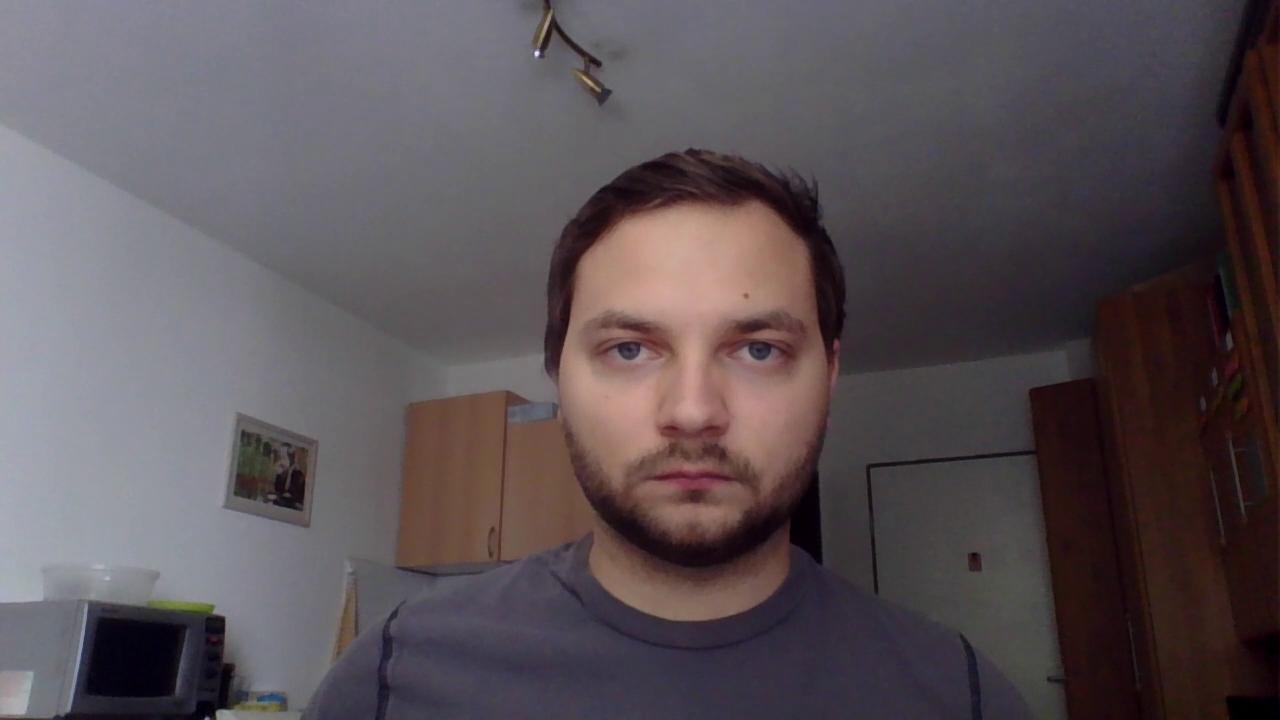}
         \\
         \includegraphics[width=\textwidth]{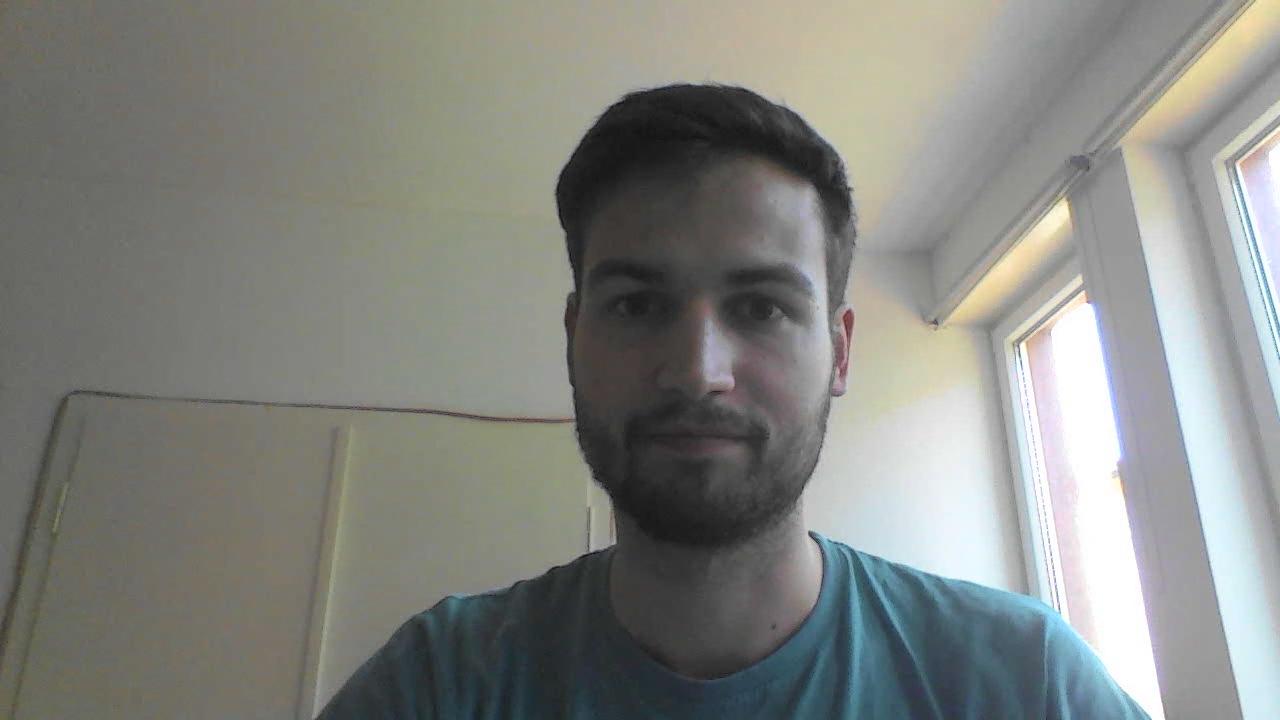}
         \caption{BL}
         \label{fig:samp:BL}
     \end{subfigure}
     \hfill
     \begin{subfigure}[b]{0.3\textwidth}
         \centering
         \includegraphics[width=\textwidth]{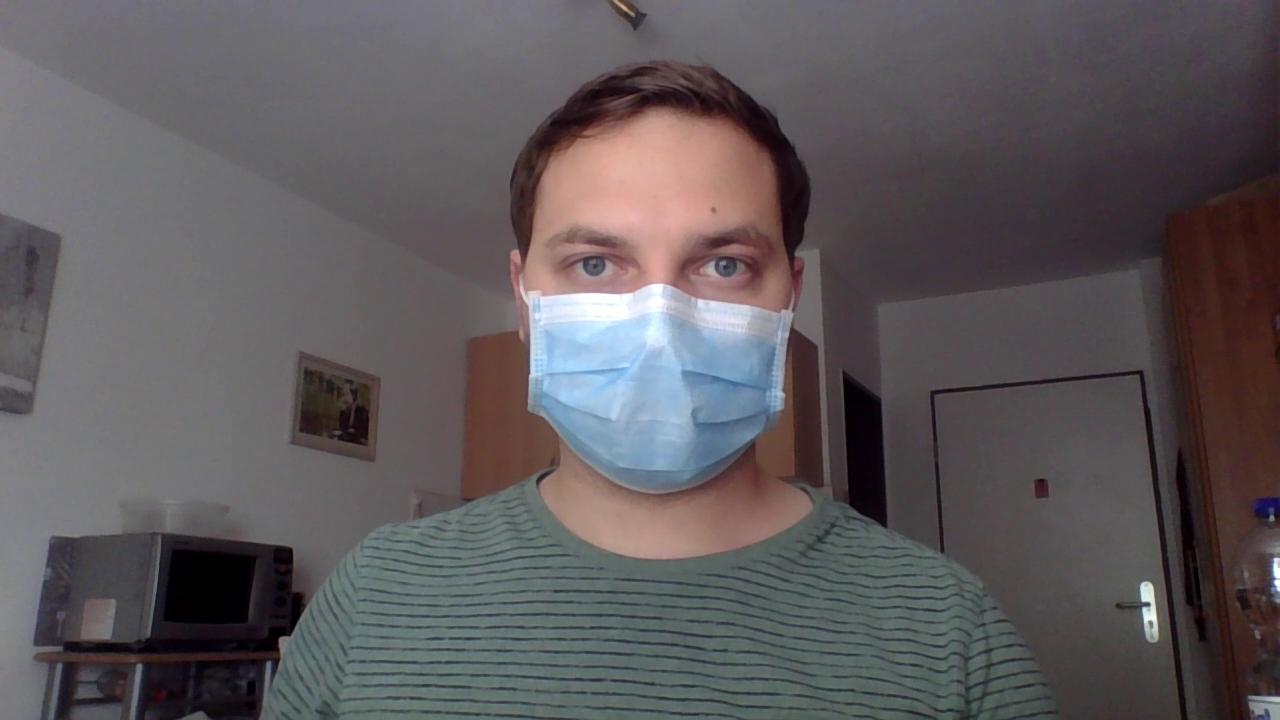}
         \includegraphics[width=\textwidth]{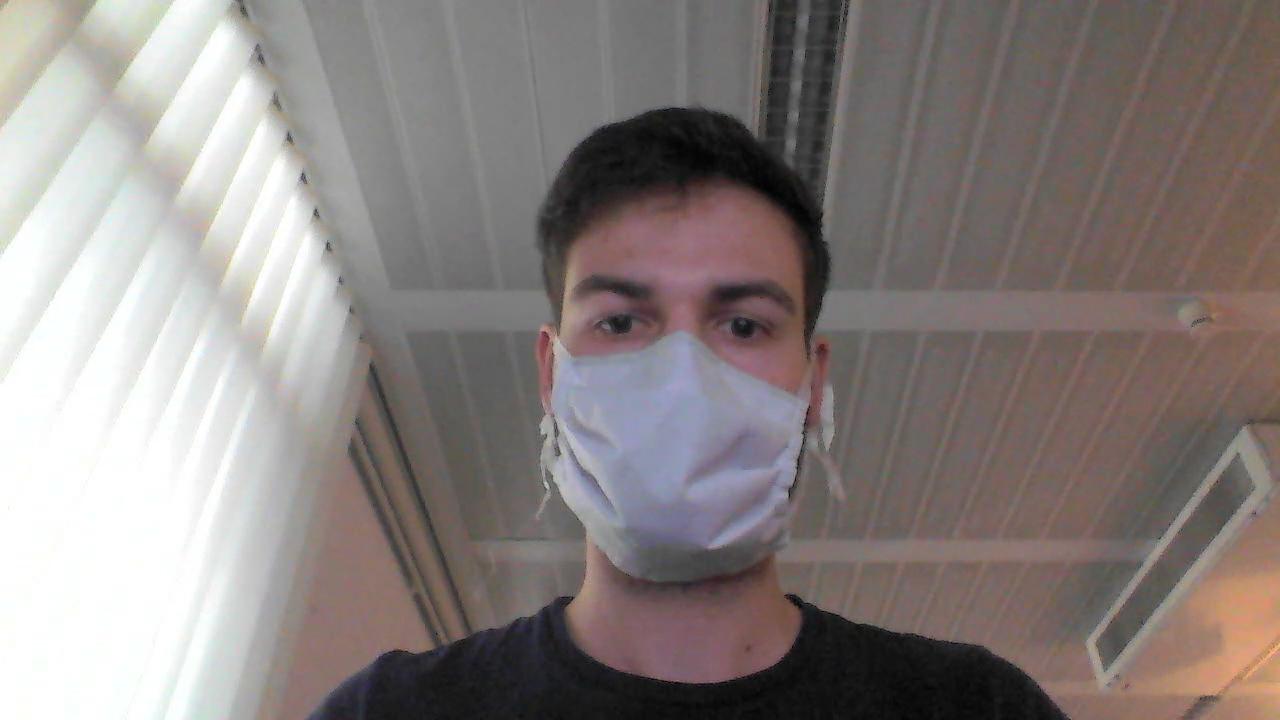}
         \caption{M1}
         \label{fig:sam:M1}
     \end{subfigure}
     \hfill
     \begin{subfigure}[b]{0.3\textwidth}
         \centering
         \includegraphics[width=\textwidth]{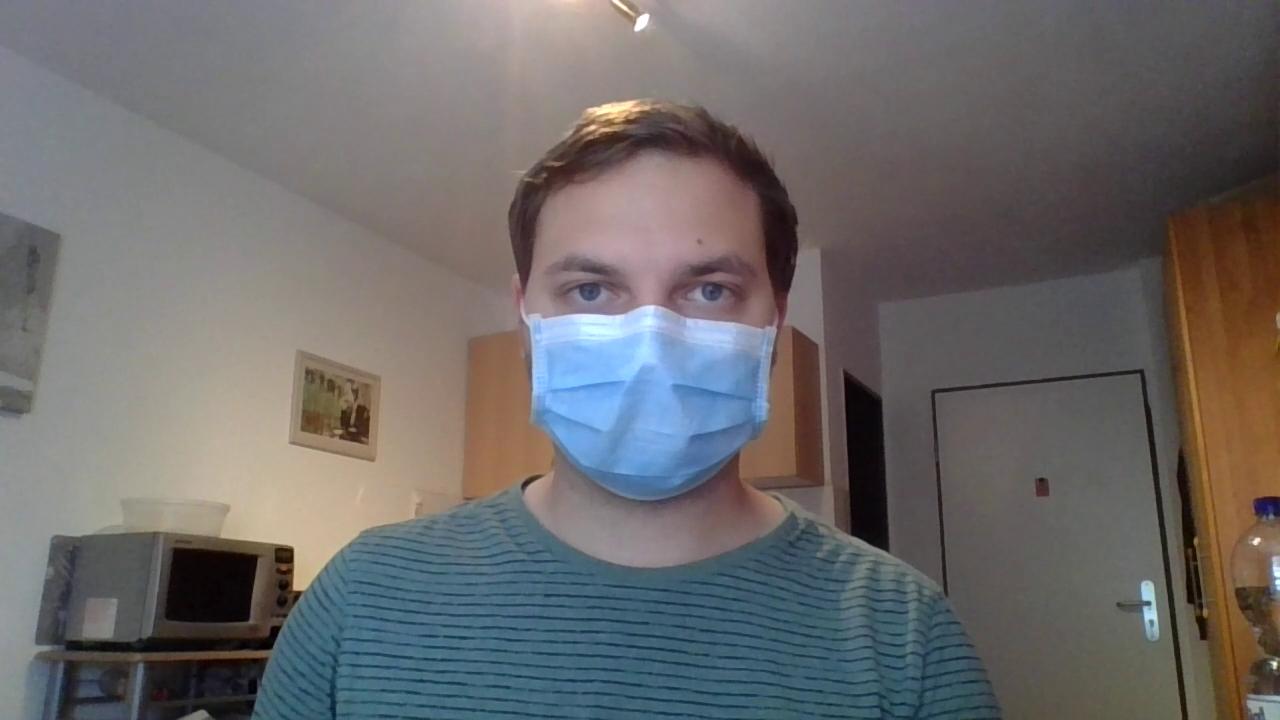}
         \includegraphics[width=\textwidth]{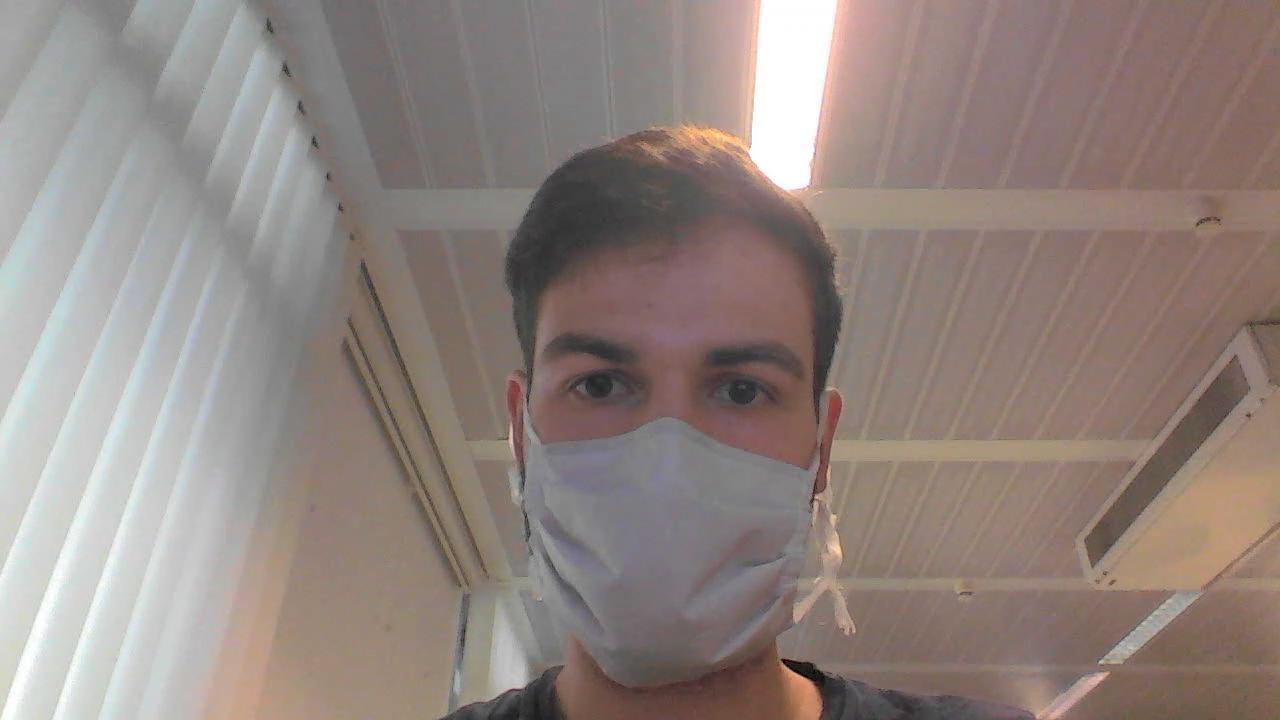}
         \caption{M2}
         \label{fig:sam:M2}
     \end{subfigure}
     
        \caption{Samples of the collected database from the three capture types (BL, M1, and M2)}
        \label{fig:samples}
				\vspace{-3mm}
\end{figure}

The first session (day) is considered as the reference data (R), resulting in the baseline reference (BLR), the mask one reference (M1R), and mask two reference (M2R). The second and third sessions (days) were considered as probe data (P) and they result in the baseline probe (BLP), the mask one probe (M1P), and mask two probe (M2P), and the joint probe data from M1P and M2P referred to as M12P. From each captured video, the first second was neglected to avoid any biases related to the user interaction with the capture device. From the following three seconds, 10 frames were selected with 9 frames gap between them, as all videos are recorded at 30 frames per second. The total number of participant at this first version of the database is 24, and they all participated in all sessions. Given the number of sessions, participants and the considered frames from each video, Table \ref{tab:DB} provide an overview on the database structure. Samples of the database are shown in Figure \ref{fig:samples}.

\begin{table}[h]
\centering
\begin{tabular}{|l|l|l|l|l|l|l|l|}
\hline
Session            & \multicolumn{3}{l|}{Session 1: References} & \multicolumn{4}{l|}{Session 2 and 3:   Probes} \\ \hline
Data split         & BLR          & M1R          & M2R          & BLP       & M1P       & M2P       & M12P       \\ \hline
Illumination       & No           & No           & Yes          & No        & No        & Yes       & Both       \\ \hline
Number of Captures & 240          & 240          & 240          & 480       & 480       & 480       & 960        \\ \hline
\end{tabular}
\caption{An overview of the database structure.}
\label{tab:DB}
\end{table}

\section{Face recognition}

To provide a wide view on the effect of wearing a mask on face recognition performance, we analyse the performance of three face recognition algorithms. Two of these algorithms are of the top performing academic approaches, namely the ArcFace \cite{DBLP:conf/cvpr/DengGXZ19} and SphereFace \cite{DBLP:conf/cvpr/LiuWYLRS17}. The third algorithm is a COTS algorithm from the vendor Neurotechnology \cite{neurotechnology}. In the following, this section provides more details on these algorithms.

\paragraph{SphereFace:}
We chose SphereFace as it achieved competitive verification accuracy on Labeled Face in the Wild (LFW) \cite{LFWTech} $99.42\%$ and Youtube Faces (YTF) \cite{wolf2011face} $95.0\%$ using 64-CNN layers trained on  CASIA-WebFace dataset \cite{DBLP:journals/corr/YiLLL14a}. SphereFace is trained using angular Softmax loss function (A-Softmax). The key idea behind  A-Softmax loss is to learn discriminative features from the face image by formulating the Softmax as angular computation between the embedded features vector $X$ and their weights $W$.

\paragraph{ArcFace:}
ArcFace achieved state-of-the-art performance of several face recognition benchmarks such as LFW $99.83\%$ and YTF $ 98.02\%$. ArcFace introduced Additive Angular Margin loss (ArcFace) to enhance the discriminative power of the face recognition model. We employed ArcFace based on ReseNet-100 \cite{DBLP:conf/cvpr/HeZRS16} architecture pretrained on refined version of MS-Celeb-1M dataset \cite{DBLP:conf/eccv/GuoZHHG16} (MS1MV2).

\paragraph{COTS:} We used the MegaMatcher 11.2 SDK \cite{neurotechnology} from the vendor Neurotechnology. We chose this COTS product as Neurotechnology achieved one of the best performances in the recent NIST report addressing the performance of vendor face verification products \cite{nist2020}. The face quality threshold was set to zero to minimize neglecting masked faces. The full processes of detecting, aligning, feature extraction, and matching are part of the COTS and thus we are not able to provide their algorithmic details. Matching two faces by the COTS produces a similarity score.

For the ArcFace \cite{DBLP:conf/cvpr/DengGXZ19} and SphereFace \cite{DBLP:conf/cvpr/LiuWYLRS17}, the Multi-task Cascaded Convolutional Networks (MTCNN) \cite{zhang2016joint} solution is used, as recommended in \cite{DBLP:conf/cvpr/LiuWYLRS17}, to detect (crop) and align (affine transformation) the face. Both network process the input aligned and cropped image and produce a feature vector of the size 512.
To compare two faces, a distance is calculated between their respective feature vectors. This is calculated as Euclidean distance for ArcFace features, as recommended in \cite{DBLP:conf/cvpr/DengGXZ19}, and as Cosine distance for SphereFace features, as recommended in \cite{DBLP:conf/cvpr/LiuWYLRS17}. The Euclidean distance (dissimilarity) is complemented to show a similarity score and the Cosine distance shows similarity score by default.



\section{Experimental setup} 

To baseline the performance, we evaluate the face verification performance without masks. This is done by N:N comparison of the data splits BLR and BLP (BLR-BLP). To measure the performance when wearing a mask, we perform an N:N comparison between the data splits BLR and M1P (BLR-M1P). To evaluate any induced performance change by having an additional illumination (room light) when wearing a mask, we perform an N:N comparison between the data splits BLR and M2P (BLR-M2P). To measure the overall performance including both considered illumination,  we perform an N:N comparison between the data splits BLR and M12P (BLR-M12P). These four experiments are used to evaluate each of the three considered face recognition solutions.

To study the effect of wearing a mask on the recognition performance, we plot the genuine and imposter distributions of the BLR-BLP (baseline) comparisons along with the genuine and imposter score distributions of the BLR-(M1P or M2P or M12P) (mask). This allows analysing the shifts in the distributions induced by wearing a mask. We also report the mean of the genuine scores (G-mean) and mean of imposter scores (I-mean) for each experiment, to get a quantitative measure of the comparison scores shifts.

Based on the standard ISO/IEC 19795-1 \cite{mansfield2006information}, we also enrich our performance study by a number of verification performance metrics. As the face mask induces a strong appearance change on the face, face detection might be challenging. Therefore, we report the failure to extract rate (FTX) for each experiment. FTX is proportion of failures of the feature extraction process to generate a template from the captures sample. Besides reporting the FTX, and only for the samples where a template can be created, we report algorithmic verification performance metrics. These metrics include the general Equal Error Rate (EER), which is defined as the common value of false mathc rate (FMR) and false non match rate (FNMR) at the decision threshold where they are identical. We also show the algorithmic verification performance by listing the FNMR at different operation points by presenting the achieved FMR100, FMR1000, and ZeroFMR, which are the lowest FNMR for an FMR $\leq$1.0\%, $\leq$0.1\%, and $\leq$0\%, respectively. To provide an algorithmic verification performance illustration on the complete range of operation points, we plots the receiver operating characteristic (ROC) curves for all the experimental setups, for each of the investigated face recognition systems.



\section{Evaluation results}

\begin{figure}[ht]
     \centering
     \begin{subfigure}[b]{0.29\textwidth}
         \centering
         \includegraphics[width=\textwidth]{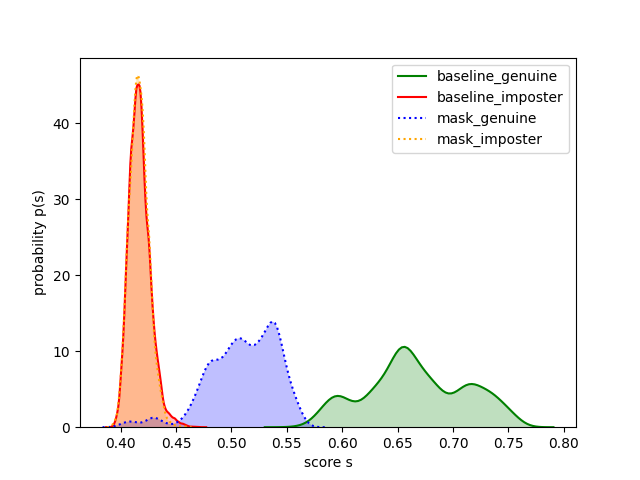}
         \caption{ArcFace: BLR-BLP and BLR-M1P}
         \label{fig:arc1}
     \end{subfigure}
     \hfill
     \begin{subfigure}[b]{0.29\textwidth}
         \centering
         \includegraphics[width=\textwidth]{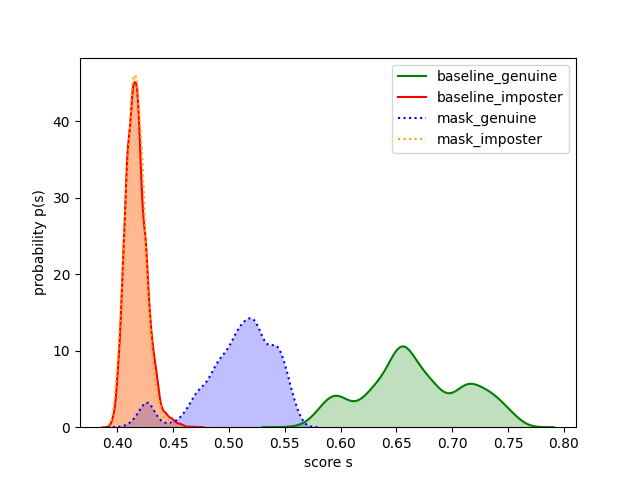}
         \caption{ArcFace: BLR-BLP and BLR-M2P}
         \label{fig:arc2}
     \end{subfigure}
     \hfill
     \begin{subfigure}[b]{0.29\textwidth}
         \centering
         \includegraphics[width=\textwidth]{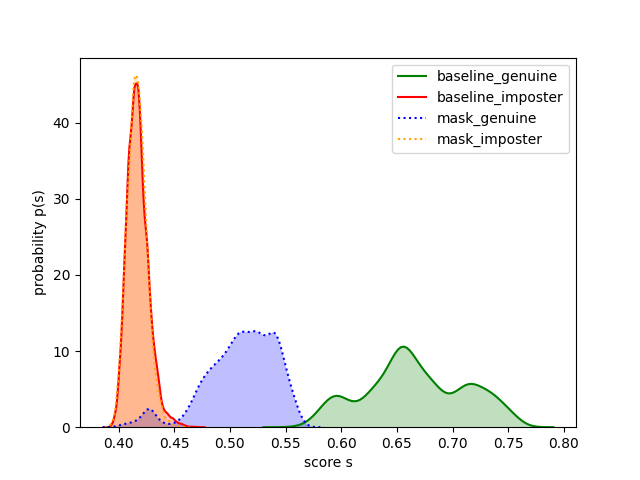}
         \caption{ArcFace: BLR-BLP and BLR-M12P}
         \label{fig:arc3}
     \end{subfigure}

     \begin{subfigure}[b]{0.29\textwidth}
         \centering
         \includegraphics[width=\textwidth]{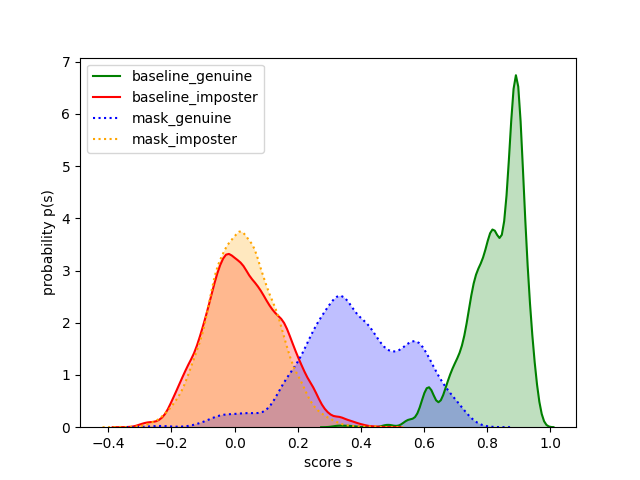}
         \caption{SphereFace: BLR-BLP and BLR-M1P}
         \label{fig:sph1}
     \end{subfigure}
     \hfill
     \begin{subfigure}[b]{0.29\textwidth}
         \centering
         \includegraphics[width=\textwidth]{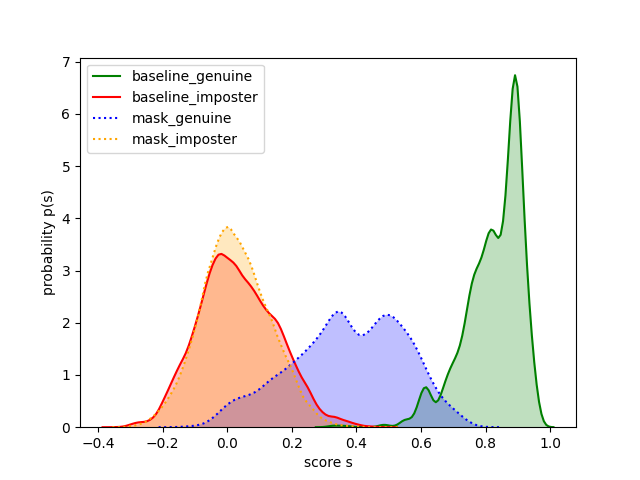}
         \caption{SphereFace: BLR-BLP and BLR-M2P}
         \label{fig:sph2}
     \end{subfigure}
     \hfill
     \begin{subfigure}[b]{0.29\textwidth}
         \centering
         \includegraphics[width=\textwidth]{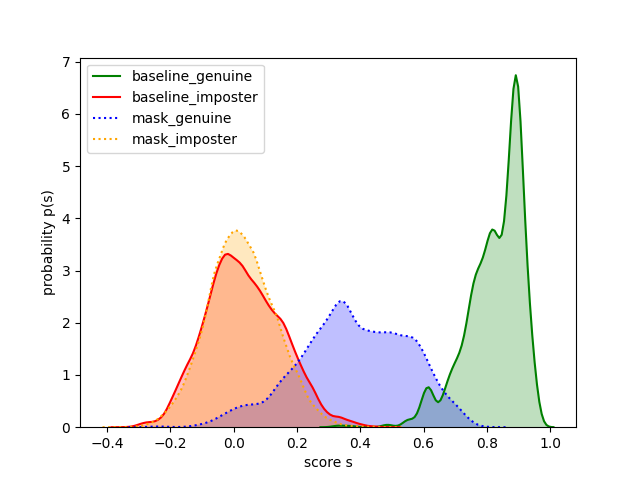}
         \caption{SphereFace: BLR-BLP and BLR-M12P}
         \label{fig:sph3}
     \end{subfigure}

     \begin{subfigure}[b]{0.29\textwidth}
         \centering
         \includegraphics[width=\textwidth]{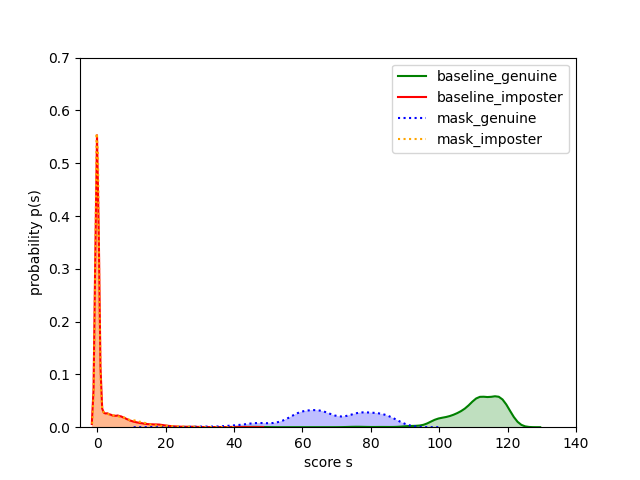}
         \caption{COTS: BLR-BLP and BLR-M1P}
         \label{fig:cots1}
     \end{subfigure}
     \hfill
     \begin{subfigure}[b]{0.29\textwidth}
         \centering
         \includegraphics[width=\textwidth]{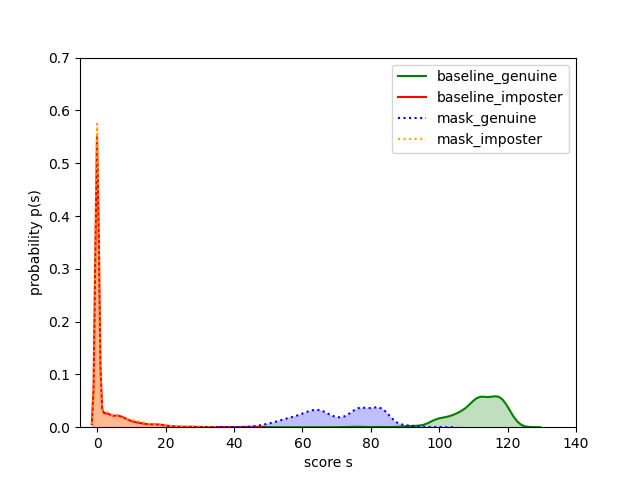}
         \caption{COTS: BLR-BLP and BLR-M2P}
         \label{fig:cots2}
     \end{subfigure}
     \hfill
     \begin{subfigure}[b]{0.29\textwidth}
         \centering
         \includegraphics[width=\textwidth]{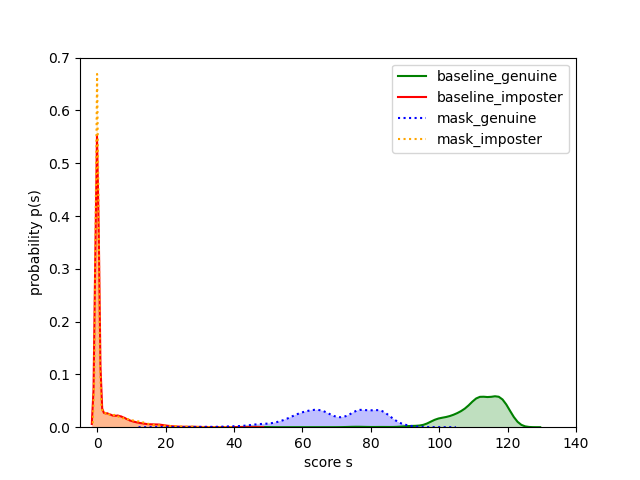}
         \caption{COTS: BLR-BLP and BLR-M12P}
         \label{fig:cots3}
     \end{subfigure}

        \caption{The comparison score (similarity) distributions comparing the "baseline" BLR-BLP genuine and imposter distributions to those of the distributions including "masked" faces probes (BLR-M1P (a, d, g)), BLR-M2P (b, e, h), BLR-M12P (c, f, i). The shift of the genuine scores towards the imposter distribution is clear when faces are masked for all investigated system (ArcFace(a, b, c), SphereFace (d, e, f), and COTS (g, h, i)).}
        \label{fig:dists}
\end{figure}

Figure \ref{fig:dists} presents the comparison between the baseline (BLR-BLP) genuine and imposter score distributions and the different masked faces experiments (BLR-M1P, BLP-M2P, BLR-M12P) on the three considered face recognition solutions. It is noticeable in all experimental setups that, when comparing masked faces probes to unmasked references, the genuine score distributions strongly shift towards the imposter distributions in comparison to the BLR-BLP setup. This indicates an expected decrease in performance and general trust in the matcher decision, as the separability between genuine and imposter samples decreases. This unwanted shift seems to be slightly stronger when the masked faces are captured under additional artificial illumination (BLR-M2P) when compared to the natural light condition (BLR-M2P). On the other hand, the imposter score distributions do not seem to be significantly affected by the masked probes (BLR-M1P, BLP-M2P, BLR-M12P) in comparison to the unmasked baseline (BLR-BLP).

Tables \ref{tab:arcMet}, \ref{tab:sphMet}, and \ref{tab:cotsMet} present the achieved performance, given by the different evaluation metrics, on all experimental setups by the ArcFace, SphereFace, and COTS solutions, respectively. In all systems, wearing a face mask affected the ability to detect the face properly, resulting in a higher than zero (as in the baseline) FTX. Interestingly, additional illumination (typically from the top) increased the FTX in all systems (BLR-M2P compared to BLR-M1P). This is probably due to the different reflection and shadow patterns induced by the illumination, see samples in Figure \ref{fig:samples}. The FTX values for the SphereFace and ArcFace in tables \ref{tab:arcMet} and \ref{tab:sphMet} are identical as they both use the MTCNN network for face detection and alignment. 

The verification performance (EER, FMR100, FMR1000, ZeroFMR) of the ArcFace and SphereFace is negatively affected when the probe faces are masked (BLR-M1P and BLR-M2P), see tables \ref{tab:arcMet} and \ref{tab:sphMet}. This negative effect is stronger when the faces are captured under the effect of artificial illumination (BLR-M2P), probably due to unexpected reflections and shadowing and the fact that the BLR references were captured without such illumination. The reduction in the performance is much more dominant in the SphereFace solution in comparison to the ArcFace. For both systems, the G-mean values decreased significantly when considering the masked probes. This, despite the small size of the evaluation data, indicates a strong negative effect of the masks on the face recognition performance. On the other hand, the I-mean value when considering the masked faces, in comparison to the baseline (BLR-BLP), was not changed under the the ArcFace solution and only slightly changed under the SphereFace solution.

\begin{table}[h]
\centering
\begin{tabular}{|l|l|l|l|l|l|l|l|}
\hline
ArcFace  & EER & FMR100 & FMR1000 & ZeroFMR & G-mean & I-mean & FTX \\ \hline
BLR-BLP  & 0.000\%   & 0.000\%      & 0.000\%       & 0.000\%       & 0.666     & 0.417      & 0.000\%   \\ \hline
BLR-M1P  & 3.163\%   & 3.517\%	      & 3.831\%       & 5.069\%       & 0.511      & 0.417      & 3.750\%   \\ \hline
BLR-M2P  & 5.504\%   & 6.163\%      & 6.628\%       & 7.616\%       & 0.509    & 0.417     & 5.833\%   \\ \hline
BLR-M12P & 4.380\%   & 4.888\%      & 5.229\%       & 6.468\%       & 0.510      & 0.417      & 4.792\%   \\ \hline
\end{tabular}
\caption{The verification performance measures, the G-mean, and I-mean achieved by ArcFace on the different experimental setups. Note the performance degradation induced by the masked face probes.}
\label{tab:arcMet}
\end{table}

\begin{table}[h]
\centering
\begin{tabular}{|l|l|l|l|l|l|l|l|}
\hline
SphereFace  & EER & FMR100 & FMR1000 & ZeroFMR & G-mean & I-mean & FTX \\ \hline
BLR-BLP  & 0.216\%   & 0.065\%      & 0.217\%       & 0.390\%       & 0.825      & 0.033      & 0.000\%   \\ \hline
BLR-M1P  & 9.312\%   & 27.35\%     & 52.95\%       & 72.91\%      & 0.384      & 0.026     & 3.750\%   \\ \hline
BLR-M2P  & 12.36\%   & 28.22\%      & 47.66\%       & 73.16\%       & 0.374    & 0.025      & 5.833\%   \\ \hline
BLR-M12P & 10.85\%   & 27.86\%      & 50.01\%      & 73.38\%       & 0.380      & 0.025      & 4.792\%   \\ \hline
\end{tabular}
\caption{The verification performance measures, the G-mean, and I-mean achieved by SphereFace on the different experimental setups. Note the performance degradation induced by the masked face probes.}
\label{tab:sphMet}
\end{table}
 
\begin{table}[h]
\centering
\begin{tabular}{|l|l|l|l|l|l|l|l|}
\hline
COTS  & EER & FMR100 & FMR1000 & ZeroFMR & G-mean & I-mean & FTX \\ \hline
BLR-BLP  & 0.249\%   & 0.000\%      & 0.251\%      & 0.668\%       & 110.8     &   2.281    & 0.000\%   \\ \hline
BLR-M1P  & 0.443\%  & 0.304\%       & 0.684\%      & 2.253\%       & 68.09       &  2.221     & 2.500\%   \\ \hline
BLR-M2P  & 0.004\%  & 0.000\%       & 0.009\%      & 0.050\%       & 70.88       &  2.298     & 3.542\%   \\ \hline
BLR-M12P & 0.239\%  & 0.152\%       & 0.341\%      & 1.237\%       & 69.49       &  2.259   & 3.021\%   \\ \hline
\end{tabular}
\caption{The verification performance measures, the G-mean, and I-mean achieved by COTS on the different experimental setups. Although the change in the performance caused by the masked face probes is insignificant, the shift in the average genuine score towards the imposter scores is very dominant in these cases.}
\label{tab:cotsMet}
\end{table}

\begin{figure}[ht]
     \centering
     \begin{subfigure}[b]{0.49\textwidth}
         \centering
         \includegraphics[width=\textwidth]{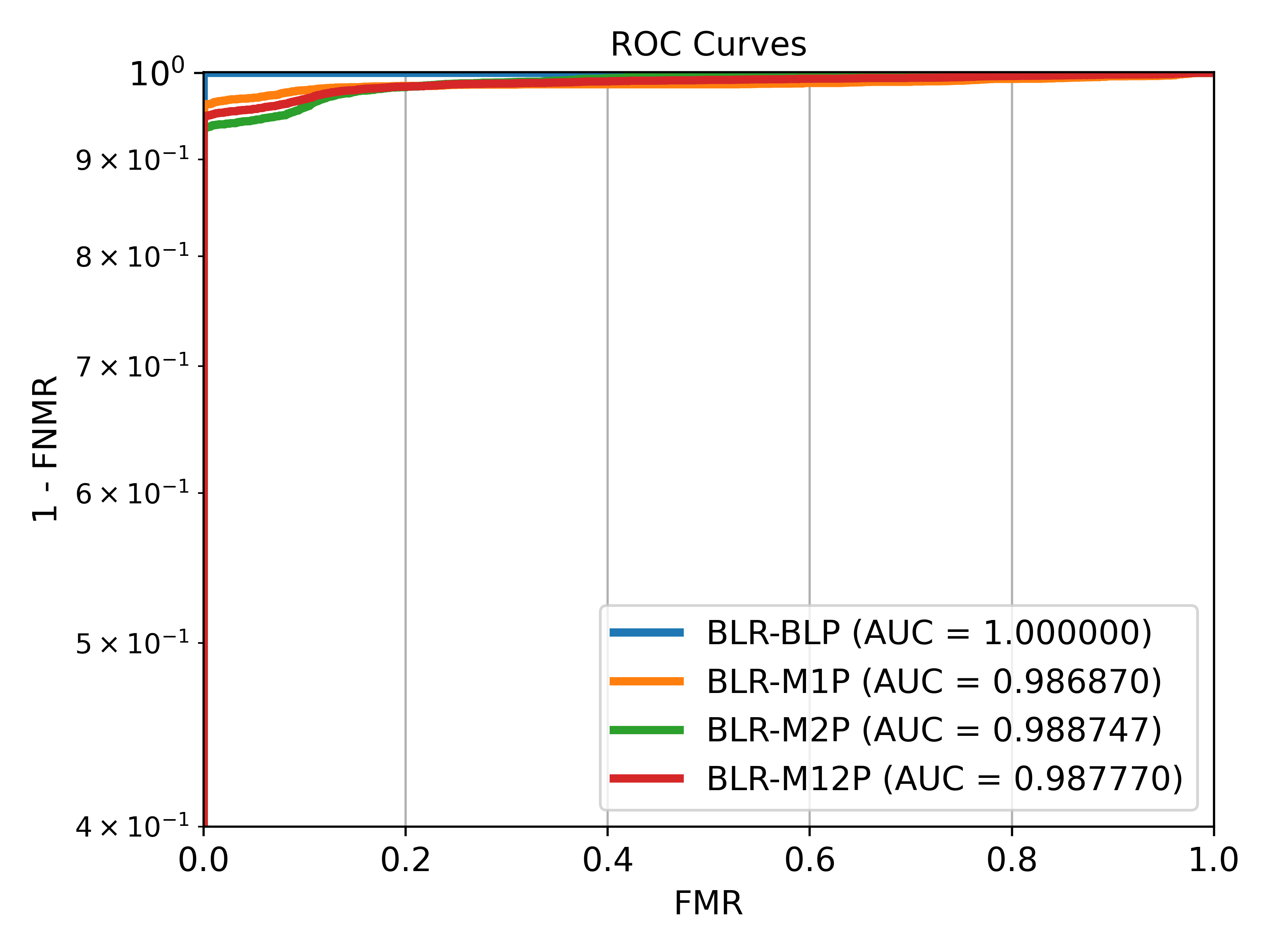}
         \caption{ArcFace}
         \label{fig:rocArc}
     \end{subfigure}
     \hfill
     \begin{subfigure}[b]{0.49\textwidth}
         \centering
         \includegraphics[width=\textwidth]{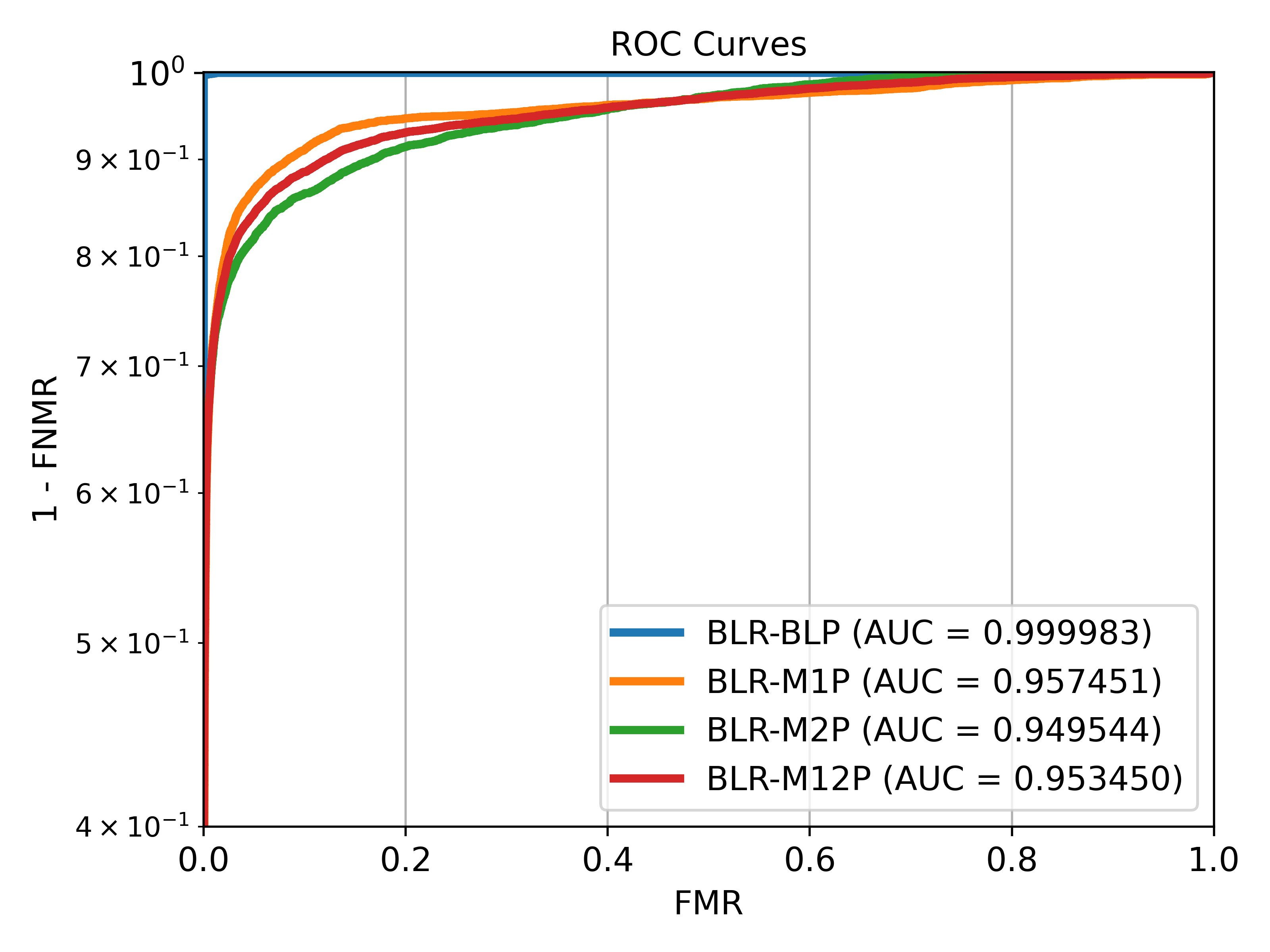}
         \caption{SphereFace}
         \label{fig:rocSphere}
     \end{subfigure}
     \hfill
     \\
     \begin{subfigure}[b]{0.49\textwidth}
         \centering
         \includegraphics[width=\textwidth]{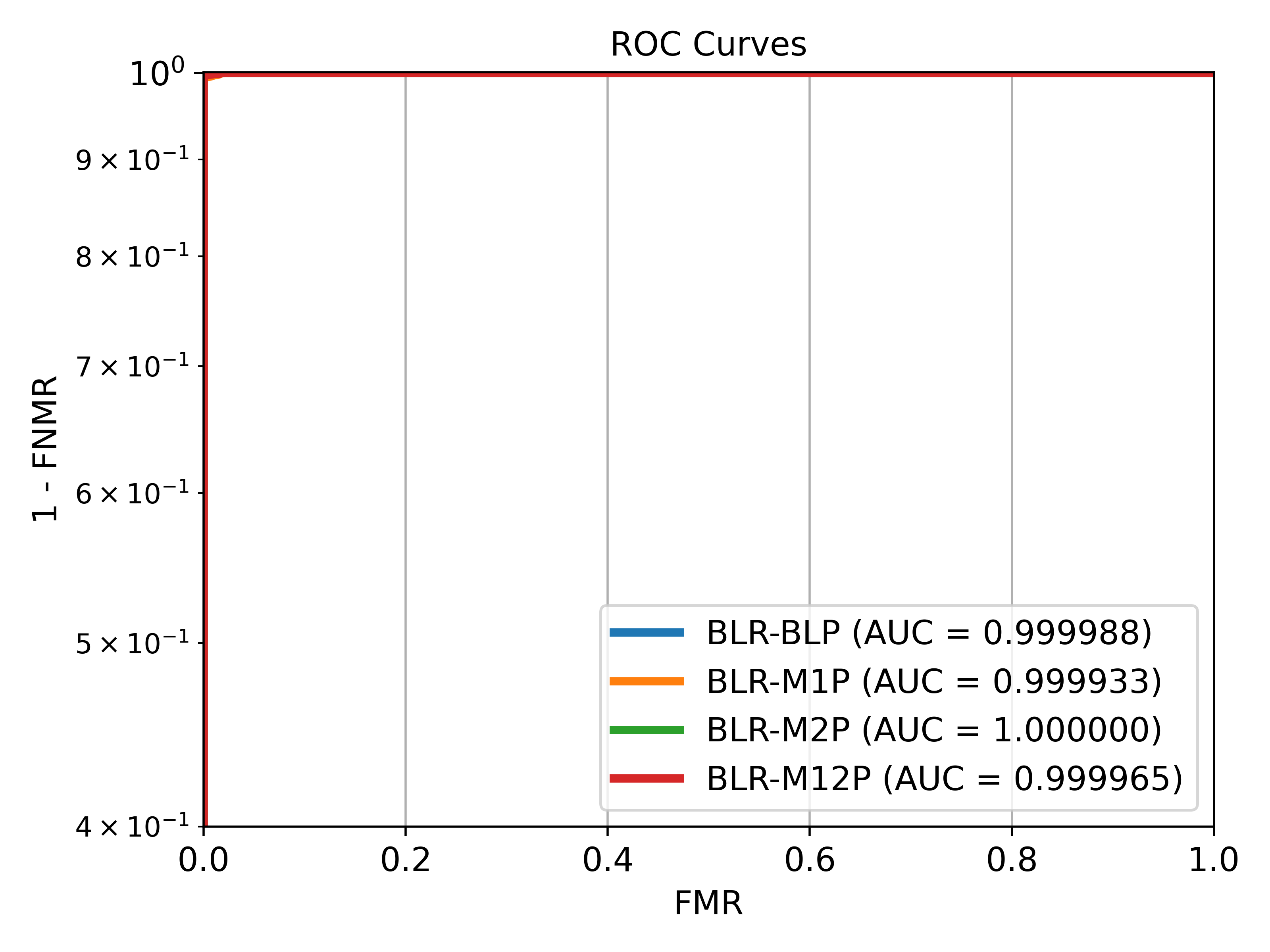}
         \caption{COTS}
         \label{fig:rocCOTS}
     \end{subfigure}
        \caption{The verification performance for the three investigated system (ArcFace(a), SphereFace (b), and COTS (c)) is presented as ROC curves. For each of the systems, four curves are plotted to represent the three settings that include "masked" faces probes (BLR-M1P, BLR-M2P, and BLR-M12P) and the unmasked baseline (BLR-BLP). The area under curve (AUC) is also listed for each of the ROC curves. As in Tables \ref{tab:arcMet}, \ref{tab:sphMet}, and \ref{tab:cotsMet}, the effect of masked probes is apparent on the performance of the ArcFace and SphereFace, while the performance of the COTS is almost perfect in all experimental settings (however, with shift in genuine scores values). }
        \label{fig:rocs}
\end{figure}

When it comes to verification performance metrics (EER, FMR100, FMR1000, ZeroFMR), the COTS is not significantly affected by masked faces. This is apparent in Table \ref{tab:cotsMet}, where these performance metrics are not significantly different in all experimental setups. This might be due to the robust and high performance of the COTS solution and the limited size of the evaluation database. However, the change in the G-mean from 110.8 in the BLR-BLP to 69.46 in the BLR-M12P, while maintaining a similar I-mean, indicates a large change in the separability (between genuine and imposter) in the COTS decisions. This can lead to an increase in the error rate given a larger and more challenging evaluation. Such an evaluation is planned as the data presented in this paper is an initial version of a larger data being collected at the moment. 
To show the verification performance over a wider range of operation points, Figure \ref{fig:rocs} presents the ROC curves for the different experimental settings for each of the three investigated systems.
Similar conclusions to those established from Tables \ref{tab:arcMet}, \ref{tab:sphMet}, and \ref{tab:cotsMet}  can be made. The ArcFace and SphereFace verification performance is effected by the masked probe faces, while the COTS maintains an almost perfect verification performance. However, one must keep in mind the significant shift in the genuine score values in all three systems, as illustrated in Figure \ref{fig:dists}.

In general, the effect of wearing face masks on the face recognition behaviour is apparent on all investigated systems. The effect is most significant on the genuine scores distribution, rather than the imposter scores distribution. This renders the current face recognition solutions undependable to match masked faces with unmasked faces and, at least, requires re-evaluation.

\section{Conclusion}

Addressing the wide spread use of face masks as a preventive measure to the COVID-19 pandemic spread, we presented an exploratory study on the effect of wearing masks on face recognition performance in collaborative scenarios.
We presented a specifically collected database captured in three different sessions, with and without wearing a mask, and is part of an ongoing effort to gather a larger scale database with realistic variations.
We analysed the behaviour of two high-performing academic face recognition solutions and one of the top performing COTS solutions.
Our analyses pointed out the significant effect of wearing a mask on comparison scores separability between genuine and imposter comparisons in all the investigated systems. Moreover, we point out a large drop in the verification performance of the academic face recognition solutions, even on a limited evaluation data, when considering masked face probes. 

\section*{Acknowledgment}
This research work has been funded by the German Federal Ministry of Education and Research and the Hessen State Ministry for Higher Education, Research and the Arts within their joint support of the National Research Center for Applied Cybersecurity ATHENE

\bibliographystyle{lnig}
\bibliography{main}

\end{document}